\documentclass[10pt]{article} 

\usepackage[preprint]{rlj} 

\usepackage{amssymb}            
\usepackage{mathtools}          
\usepackage{mathrsfs}           
\usepackage{graphicx}           
\usepackage{subcaption}         
\usepackage[space]{grffile}     
\usepackage{url}                
\usepackage{lipsum}             
\usepackage{seqsplit}
\usepackage{booktabs}
\usepackage{tabularx}
\usepackage{array}

\newcommand{\repoURL}{\url{https://github.com/taodav/jaxenstein}}
\newcommand{\baselinesURL}{\url{https://github.com/taodav/jaxenstein_baselines/}}

\newcommand{\envID}[1]{\texttt{\seqsplit{#1}}}

\title{JAXenstein: Accelerated Benchmarking for First-Person Environments}

\setrunningtitle{JAXenstein: Accelerated Benchmarking for First-Person Environments}

\author{Ruo Yu Tao, George Konidaris}

\emails{ruo\_yu\_tao@brown.edu}

\affiliations{
\textbf{Department of Computer Science, Brown University, USA}\\
}


\begin{document}

\maketitle  

\begin{abstract}
The progression of reinforcement learning algorithms have been driven by challenging benchmarks.
The rate in which a researcher can iterate on a problem setting directly impacts the speed of algorithm development.
Modern machine learning has produced tools that allow for fast and scalable algorithm development like the JAX library.
With the availability of these tools, a serious bottleneck in algorithm development is the availability of large and complex domains for experimentation.
Most notably, the JAX reinforcement learning ecosystem does not have any benchmarks that test visual first-person tasks; these domains are crucial for testing both exploration and an agent's ability to overcome partial observability.
We introduce JAXenstein\footnote{Code: \repoURL\\\indent\hspace{5px} Baselines: \baselinesURL} an open-source JAX-based benchmark that implements the Wolfenstein 3D rendering engine for fast and scalable experimentation in visual first-person tasks.
JAXenstein is several times faster than comparable vision-based benchmarks, and is easily extensible to more complex first-person domains.
\end{abstract}
\section{Introduction}
\label{sec:intro}

The development of reinforcement learning algorithms have been driven by challenging benchmarks. 
Problem settings have preceded the development of algorithms, as they act as a foundation for algorithmic improvement.
Most famously, the arcade learning environment~\citep{bellemare13arcade} gave rise to deep Q-networks~\citep{mnih2015humanlevel} and discrete, deep value-based control, while the Mujoco benchmark~\citep{todorov2012mujoco} was the testbed for deep continuous control and policy-gradient methods~\citep{silver2014dpg,lilicrap2016ddpg,schulman2017ppo}. 

The speed of benchmarks directly affect the rate of experimentation and hence the speed of algorithmic development. 
A reinforcement learning research project will usually start with a faster, more lightweight benchmark, such as Cart Pole~\citep{barto1983neuron} or Mountain Car~\citep{moore90efficientmemory}, to develop an algorithm before moving on to larger domains which require more time to train on. Historically, these larger domains requires much time to train algorithms. With the introduction of multiprocessing~\citep{mnih2016a3c} and GPU accelerated~\citep{brax2021github} environments, the speed, efficiency and parallelizability of environments has made iterating on larger domains more accessible, albeit at the cost of more computation. 

With the explosion in modern deep learning tools, a key bottleneck for algorithm development is the need for scalable, larger, more complex domains. 
Reinforcement learning experimentation has seen large speed and parallelization gains due to the adoption of the JAX~\citep{jax2018github} library, which can run experiments several times faster than one trained with traditional frameworks~\citep{lu2022discovered}. This requires both the algorithm and environment be entirely implemented in this framework, which means benchmarks not implemented in JAX are unavailable or costly. Notably, first-person domains~\citep{beattie2016dmlab,Wydmuch2019ViZDoom} that were originally introduced as candidate challenge problems have not been widely adopted in JAX frameworks due to the complexity in reimplementing rendering engines. These benchmarks are crucial for testing decision making in visual domains that both need to overcome partial observability and require good exploration.



We introduce JAXenstein: a lightweight JAX-based~\citep{jax2018github} implementation of the Wolfeinstein 3D engine for accelerated first-person vision-based environments. 
JAXenstein environments are simplified versions of first-person vision-based environments used in reinforcement learning benchmarking~\citep{Wydmuch2019ViZDoom,beattie2016dmlab}, foregoing visual rendering assets for speed. 
JAXenstein is entirely implemented in JAX with just-in-time (\texttt{jit}) compilation and vectorized mapping (\texttt{vmap}) support, and runs at $\sim6\times$ the speed of the ViZDoom simulator on a single CPU core. Together with a purely end-to-end JAX pipeline, JAXenstein allows for fast, GPU-accelerated training many times faster than a comparable implementation with PyTorch.
\begin{figure}[t]
    \centering
    \begin{subfigure}[b]{0.32\textwidth}
        \centering
        \includegraphics[width=\textwidth]{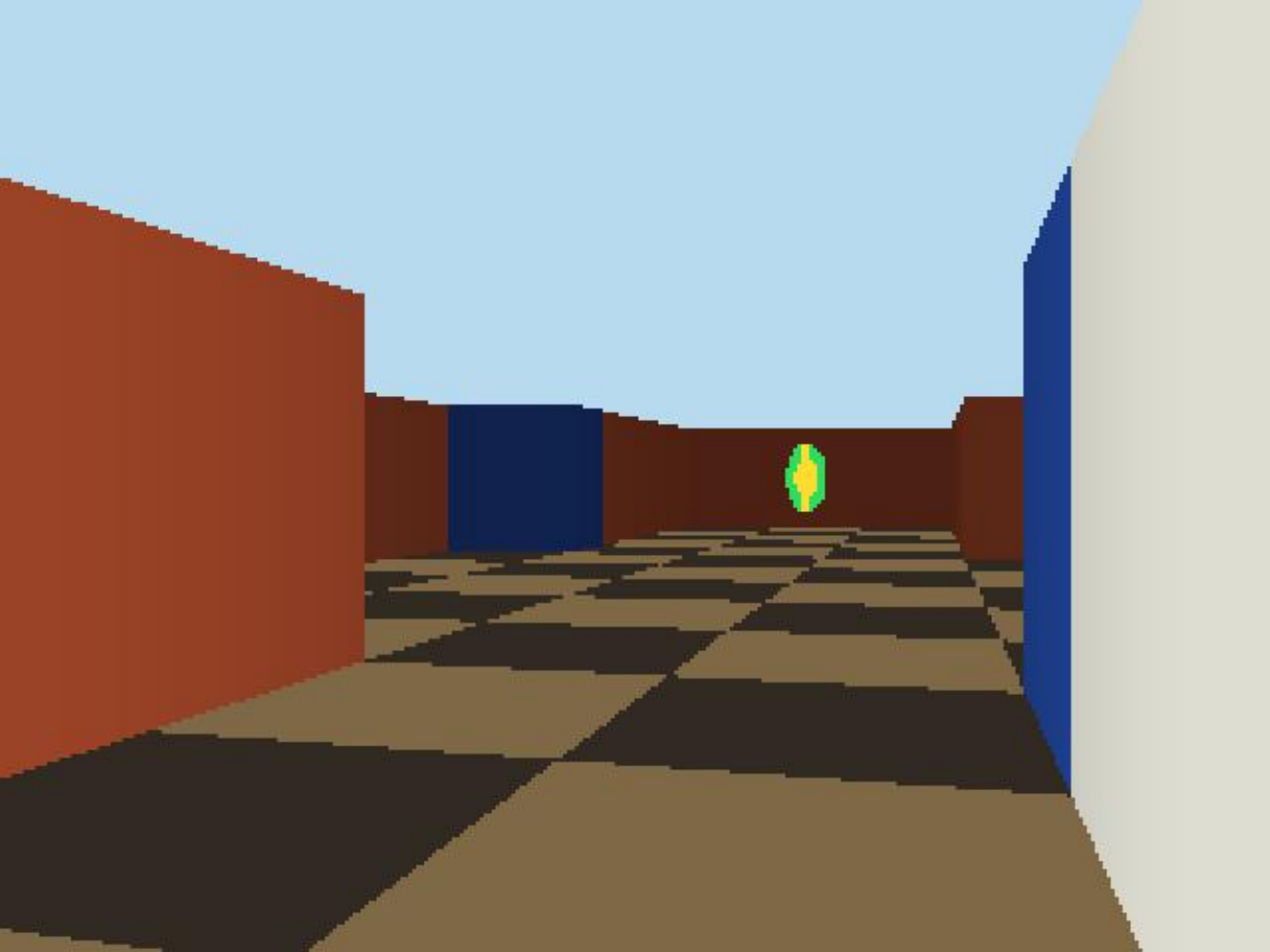}
        \caption{\texttt{my-way-home}}
        \label{fig:my-way-home}
    \end{subfigure}
    \hfill
    \begin{subfigure}[b]{0.32\textwidth}
        \centering
        \includegraphics[width=\textwidth]{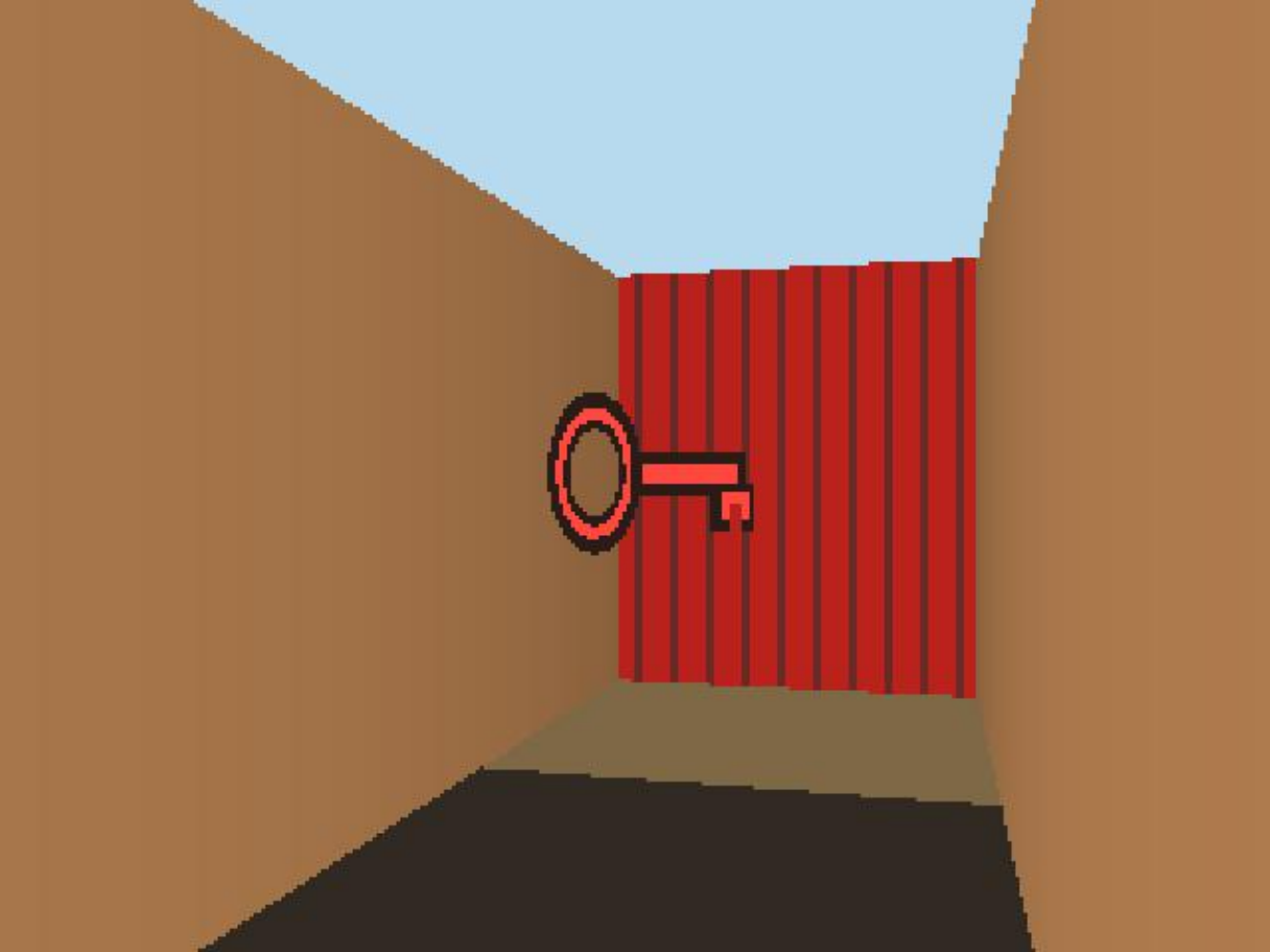}
        \caption{\texttt{key-door}}
        \label{fig:key-door}
    \end{subfigure}
    \hfill
    \begin{subfigure}[b]{0.32\textwidth}
        \centering
        \includegraphics[width=\textwidth]{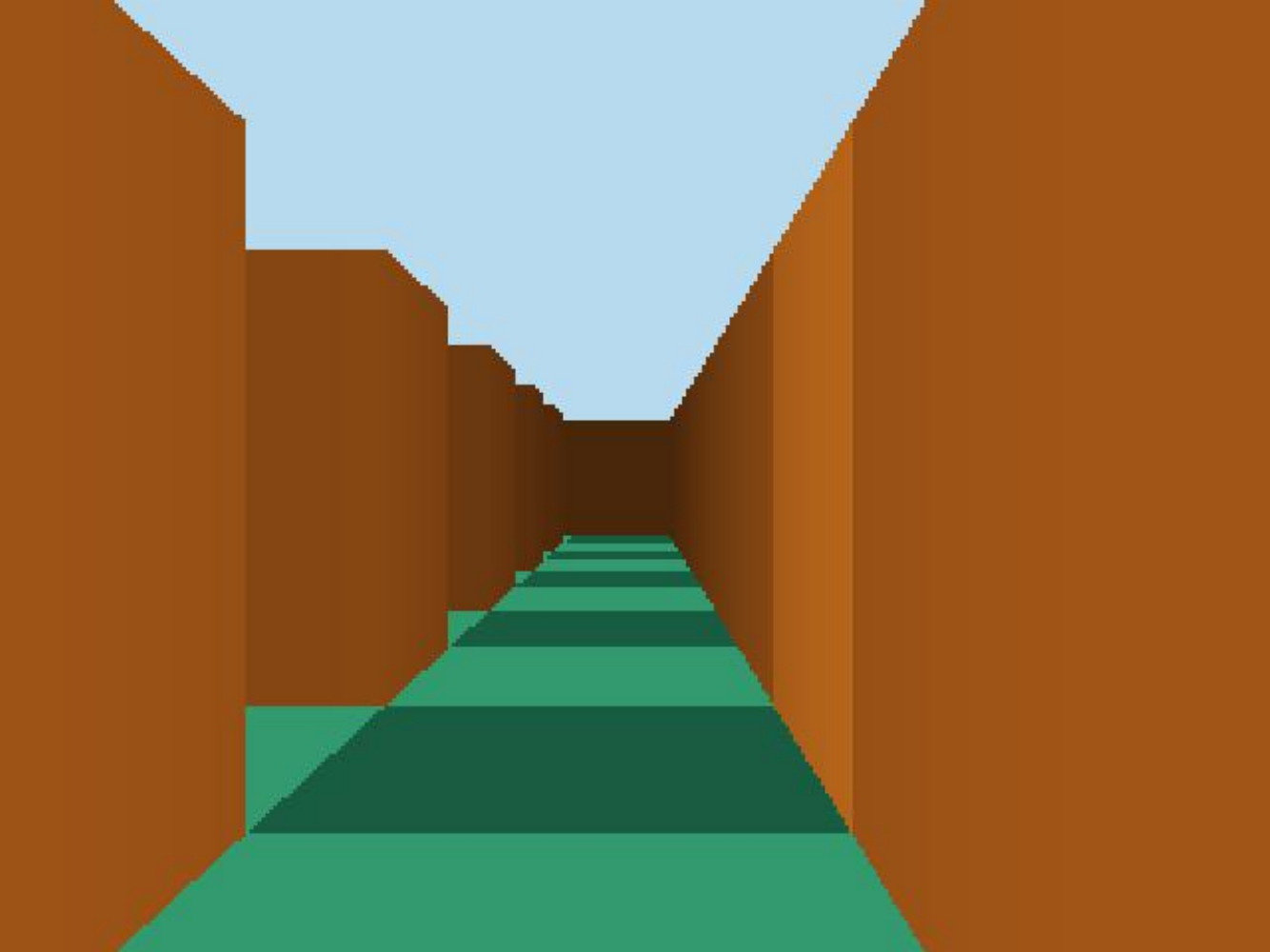}
        \caption{\texttt{dmlab-static-03}}
        \label{fig:dmlab-02}
    \end{subfigure}
    \caption{Example environments in the JAXenstein benchmark, entirely implemented in end-to-end JAX. Objects are rendered as fixed sprites, like in the original Wolfenstein 3D engine.}
    \label{fig:JAXenstein_stills}
\end{figure}

\section{A simple JAX-native rendering engine}
\label{sec:JAXenstein}
JAXenstein runs on a simple rendering engine that uses ray casting, first introduced in a production game in Wolfenstein 3D~\citep{idsoftware1992wolfenstein3d}. JAXenstein uses this simple 3D rendering engine because its speed: ray casting is extremely fast in rendering frames for simple 3D environments based on tile maps, allowing for fast and scalable environment interactions. We show a few examples of this rendering in Figure~\ref{fig:JAXenstein_stills}.
We use this front-end as a rendering engine to reimplement light and fast versions of existing first-person reinforcement learning benchmarks.

\subsection{Ray casting}

Ray casting produces the scene of an image from an observer by calculating, from each column in an image, the intersect of a ray cast from the observer to an intersection in the scene (e.g. a wall). 
We can use an appropriate shading color to give depth to the rasterization based on the distance of this intersection.

We calculate this intersection with the digital differential analyzer (DDA) algorithm~\citep{watt20003d}. The DDA algorithm leverages the fact that we are in a tile-based world, and finds the intersection for a particular ray by ``marching'' along the tile boundaries that intersect with the ray. This requires far fewer steps in the ray casting algorithm as compared to marching along the ray in a set, small interval, and results in significantly fewer intersection checks.

\subsection{Performance trade-offs}

While ray casting with DDA is an extremely fast and simple algorithm for rendering, it does incur trade-offs.
First and most importantly, it requires that the environment is a tile-based map. While fitting for many simple first-person environments, anything with more complex environment objects needs more complex algorithms that would necessarily slow down the simulator.
This also implies that stairs, jumping, height differences are not possible to render with this engine. 
While extensions are possible for more complex objects~\citep{mordvintsev2022jaxraytracing}, this comes at a computational cost that will affect per-timestep performance.

Nonetheless, this rendering engine is the first to introduce first-person environments into the JAX reinforcement learning ecosystem. Using this renderer, we port existing first-person environments used in reinforcement learning benchmarks for extremely fast and scalable experimentation.

\section{JAXenstein: a benchmark suite for accelerated first-person environments}

\begin{figure}[t]
    \centering
    \begin{subfigure}[t]{0.43\textwidth}
        \centering
        \includegraphics[width=\textwidth]{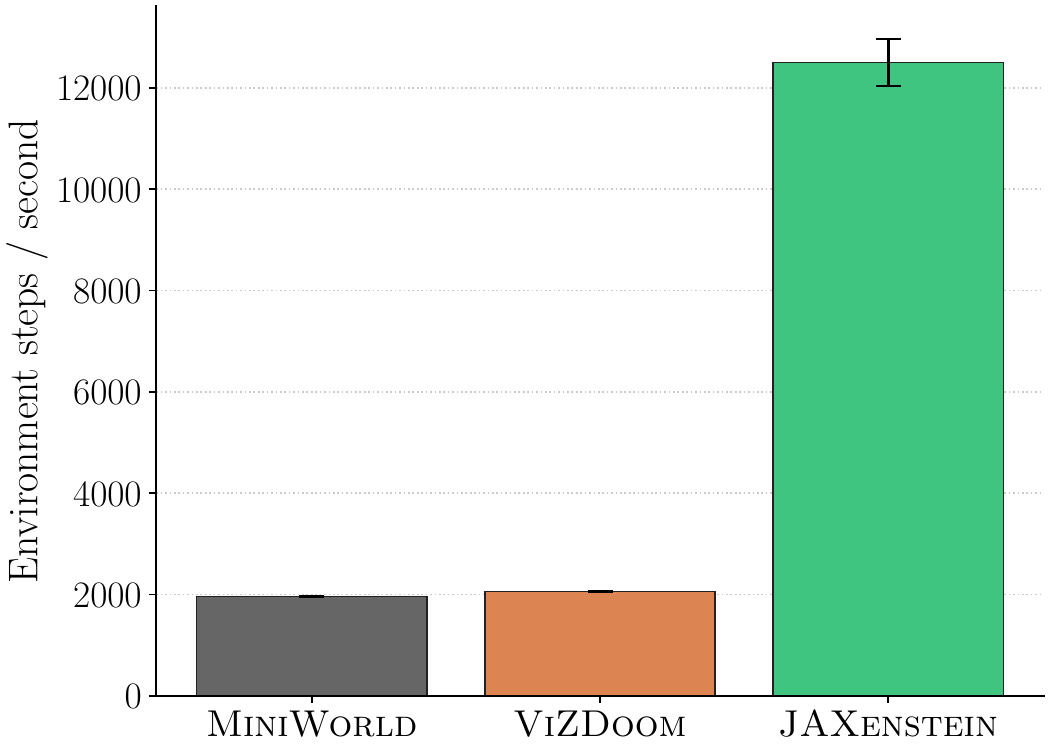}
        \caption{Steps-per-second comparison. JAXenstein is approximately $6\times$ the speed of MiniWorld and ViZDoom.}
        \label{fig:env-timing}
    \end{subfigure}
    \hfill
    \begin{subfigure}[t]{0.54\textwidth}
        \centering
        \includegraphics[width=\textwidth]{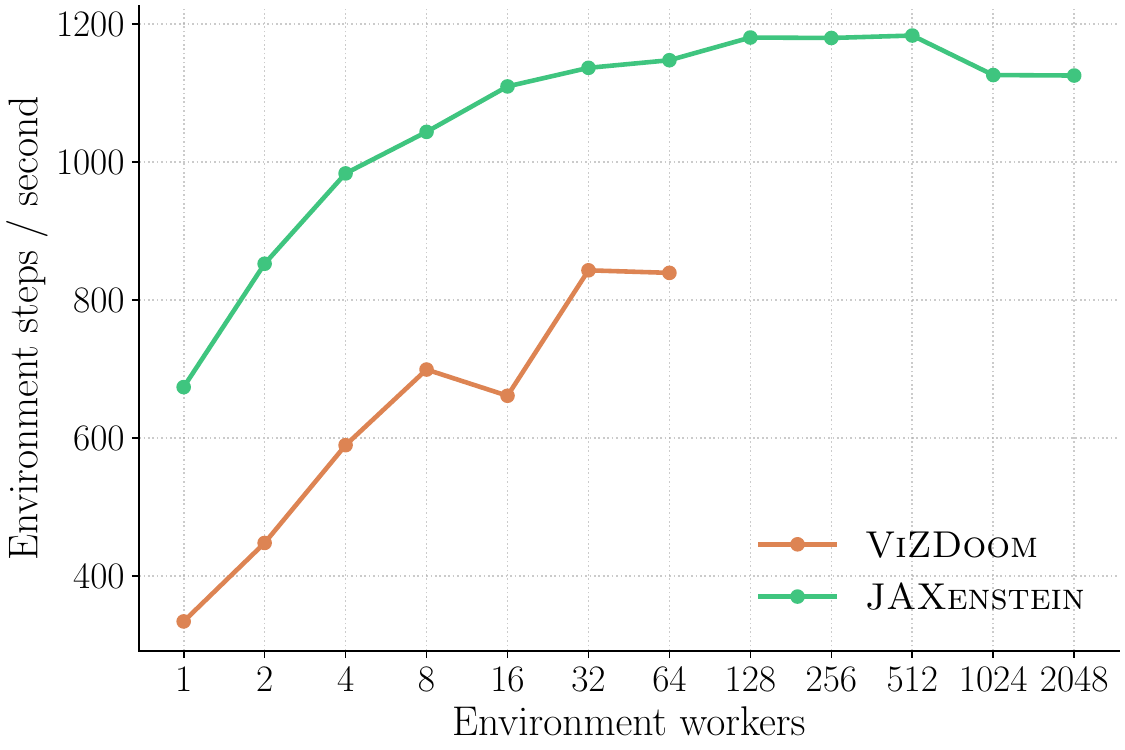}
        \caption{Training speed comparison between ViZDoom and JAXenstein. ViZDoom crashes after trying to parallelize > 64 environments.}
        \label{fig:train-timing}
    \end{subfigure}
    \caption{Speed comparisons between JAXenstein and similar benchmarks.}
    \label{fig:runtime}
\end{figure}

We introduce JAXenstein: a purely JAX-based benchmark for first-person environments based on the Wolfenstein 3D engine. This benchmark includes simplified reimplementations of popular first-person environments used throughout reinforcement learning research, such as ViZDoom~\citep{Wydmuch2019ViZDoom} and DeepMind Lab~\citep{beattie2016dmlab}, as well as completely new environments.

\subsection{Runtime comparison}
We conduct experiments to test the computational efficiency of JAXenstein compared to other benchmarks.
We first compare JAXenstein run times to MiniWorld~\citep{MinigridMiniworld23} and ViZDoom in Figure~\ref{fig:env-timing}. JAXenstein achieves around a $6\times$ speed up in steps per second. Details of these runs are given in Section~\ref{appx:runtime_comparison}.

One reason JAXenstein achieves comparably faster speeds is due to the availability of native rendering at lower resolutions. Since images are usually downsampled to a smaller size in deep reinforcement learning algorithms, rendering at this smaller size allows for massive per-time-step speed ups, without the per-timestep cost of rescaling images.

We also compare a training time comparison when scaling up the number of parallel environments in Figure~\ref{fig:train-timing}. Using a base recurrent proximal policy optimization (PPO)~\citep{schulman2017ppo} algorithm over 1M steps, compare a pure JAX end-to-end training pipeline~\citep{lu2022discovered} compared to ViZDoom with a Stable Baselines3~\citep{stable-baselines3} PPO implementation, the JAXenstein implementation achieves much higher environment steps per second. This, together with the ability for a pure JAX training loop to utilize \texttt{vmap} allows for both massive scaling and extremely fast experimentation with GPU acceleration. Training experiment details are in Section~\ref{appx:train_comparison}.

\subsection{Environments}
The ray casting engine affords many options for first-person environments. Since DDA ray casting uses a tile-based representation of the environment, this allows for a variety of options for reinforcement learning tasks, all with a common map definition format.

\paragraph{ASCII mazes}
JAXenstein includes functionality to convert any 2D maze defined over ASCII characters. We introduce the \texttt{simple} and \texttt{key-door} domains, which are simple navigation environments. JAXENSTSEIN allows for conversion of any ASCII maze into a first-person environment. All that is required to define an environment is to define a multi-line ASCII string which represents the map based on differing characters which map to elements in the map. A full definition of all characters are described in Section~\ref{appx:ascii_maze_details}.
Additionally, using this ASCII mapping we reimplement the MiniGrid \texttt{MiniGrid-KeyCorridorS4R3-v0} environment in JAXenstein as \texttt{key-corridor}.

\paragraph{ViZDoom environments} JAXenstein also includes its own version of ViZDoom~\citep{Wydmuch2019ViZDoom} environments. These include the My Way Home (\texttt{my-way-home}) and Health Gathering (\texttt{health-gathering}) environments.
Environments that involve shooting and enemy interaction are also possible with the JAXenstein engine, but will be added in a later release.

\paragraph{DeepMind Lab navigation mazes} To test hard first-person navigation environments, we port over the navigation domains from the DeepMind Lab benchmark~\citep{beattie2016dmlab}. These environments are over three maze configurations, with options \texttt{static} or \texttt{random} to denote whether the goal state is static or randomly sampled at each new episode. There are three maps (indexed \texttt{01}, \texttt{02} and \texttt{03}) of increasing size and complexity in this domain, giving us a total of six navigation mazes ported over from DeepMind Lab.

All implemented environments are listed in Table~\ref{tab:environments}.

\begin{table}[ht]
\centering
\small
\setlength{\tabcolsep}{3pt}
\renewcommand{\arraystretch}{1.15}
\begin{tabularx}{\columnwidth}{@{}p{0.13\columnwidth}p{0.20\columnwidth}p{0.24\columnwidth}X@{}}
\toprule
Group & Environment & ID & Description \\
\midrule
Basic & Simple & \envID{simple} & Small navigation task with one start and one goal. \\
Basic & Key-door & \envID{key-door} & Collect a red key, open a red door, reach the goal. \\
MiniGrid & KeyCorridorS4R3 & \envID{key-corridor} & $3\times3$ room key corridor with colored doors and a locked goal room. \\
ViZDoom & Health Gathering & \envID{health-gathering} & Survive an acidic room by collecting medkits. \\
ViZDoom & My Way Home & \envID{my-way-home} & Large maze with many starts, colored walls, and one goal. \\
DMLab & Static goal & \envID{dmlab-static-\{01,02,03\}} & Fixed-goal mazes; \texttt{01} small, \texttt{02} medium, \texttt{03} large. \\
DMLab & Random goal & \envID{dmlab-random-goal-\{01,02,03\}} & Same sizes; one randomly sampled goal candidate is active each episode. \\
\bottomrule
\end{tabularx}
\caption{JAXenstein environments.}
\label{tab:environments}
\end{table}

\subsection{Baseline results}
\begin{figure}[t]
    \centering
    \begin{subfigure}[t]{0.32\textwidth}
        \centering
        \includegraphics[width=\textwidth]{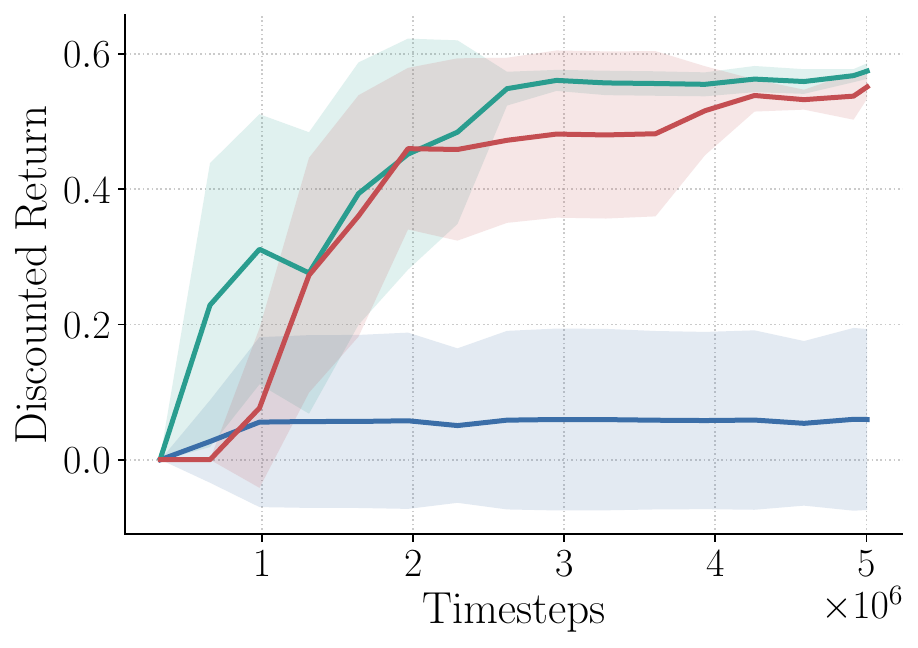}
        \caption{\texttt{simple}}
        \label{fig:simple_results}
    \end{subfigure}
    \hfill
    \begin{subfigure}[t]{0.32\textwidth}
        \centering
        \includegraphics[width=\textwidth]{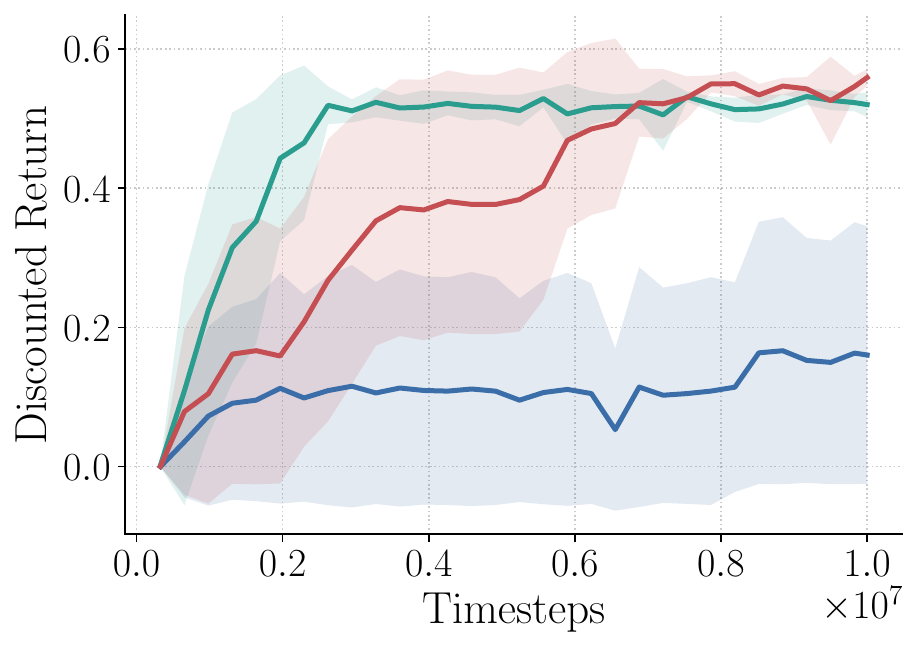}
        \caption{\texttt{key-door}}
        \label{fig:key_door_results}
    \end{subfigure}
    \hfill
    \begin{subfigure}[t]{0.32\textwidth}
        \centering
        \includegraphics[width=\textwidth]{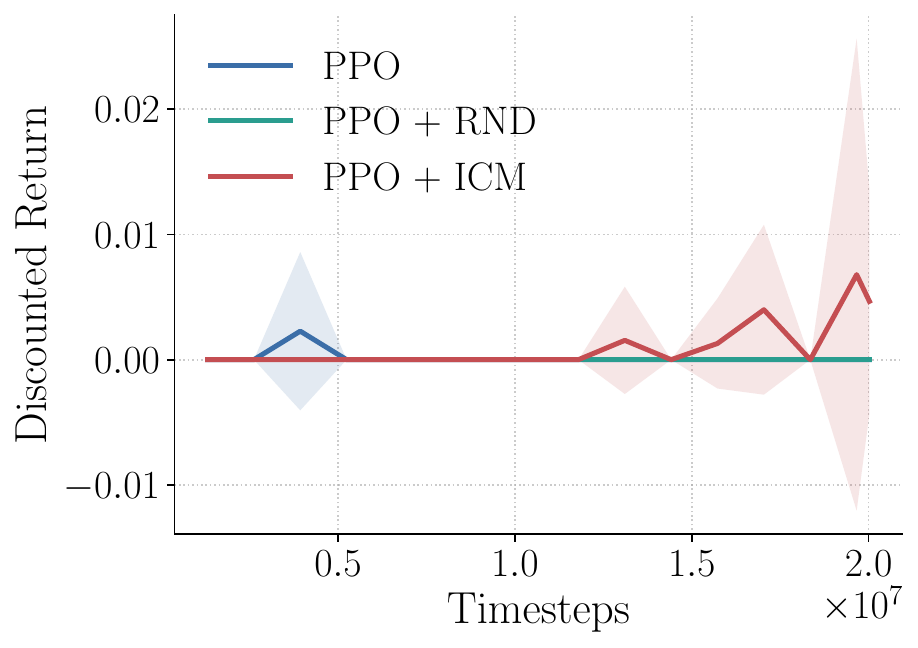}
        \caption{\texttt{dmlab-static-01}}
        \label{fig:dmlab_static_01_results}
    \end{subfigure}
    \caption{Baseline results across JAXenstein environments. Runs were conducted on the recurrent PPO algorithm with different exploration strategies. Full experimental details in Section~\ref{appx:baseline_results}.}
    \label{fig:baselines}
\end{figure}

We run experiments with the recurrent PPO~\citep{schulman2017ppo} algorithm augmented with different exploration algorithms in Figure~\ref{fig:baselines}. Experiments were conducted on a selection of the environments available in the JAXenstein benchmark. \texttt{simple} and \texttt{key-door} were included as easy sanity-checking environments. Interestingly, performance of these three algorithms are all poor in the \texttt{dmlab-static-01} environment, which is the easiest maze navigation environment among all the benchmarks. This reveals the need for better memory mechanisms and exploration methods for first-person environments.

Environments in JAXenstein require both recurrency and exploration to solve tasks. Due to the large and complex observation space, and the partial observability of the environment, JAXenstein is a benchmark that tests the memory capabilities of an algorithm, as well as the exploration capabilities when dealing with partial observability. Even in the relatively small \texttt{simple} environment, an agent that does not leverage good exploration strategies are not able to learn the task. More involved exploration algorithms like random network distillation~\citep{burda2018exploration} and intrinsic curiosity module~\citep{pathak2017curiosity} are required to solve this simple task due to its large and complex state/observation space.

\section{Conclusion}
We present JAXenstein, a fast and scalable visual first-person benchmark based on the Wolfenstein 3D rendering engine. JAXenstein is entirely implemented in JAX, allowing for fast GPU accelerated experimentation. JAXenstein reimplements a lightweight version of existing visual first-person environments from existing benchmarks like ViZDoom and DeepMind Lab, and also converts 2D gridworld environments into first-person. In the future, we hope to implement more functionality to the environment that would allow for more complex environments, like adding non-playable characters and movable objects. 









\bibliography{main}
\bibliographystyle{rlj}

\beginSupplementaryMaterials

\section{Runtime comparison details}
\label{appx:runtime_comparison}
All runtime comparisons were done on a single desktop with AMD Ryzen 9 5900X process with 32GB of system memory, and an NVIDIA RTX3090 with 24GB of VRAM.

\subsection{Steps per second}
We detail the runtime experiments in Figure~\ref{fig:env-timing}. A uniform random policy was unrolled for all 3 benchmark environments. JAXenstein ran the \texttt{my-way-home} environment, ViZDoom ran the \texttt{ViZDoomMyWayHome-v1} environment, whereas MiniWorld ran the \texttt{MiniWorld-Hallway-v0} environment. While the large discrepancy for MiniWorld may seem an issue, we note that the environment ran for this library was much simpler than the My Way Home environment, and should be much faster to run than a comparable version of My Way Home with the same engine, since the number of rooms is substantially lower. This environment should be considered an upper-bound for environment steps per second on the MiniWorld benchmark.

\begin{figure}[h]
    \centering
    \includegraphics[width=0.75\linewidth]{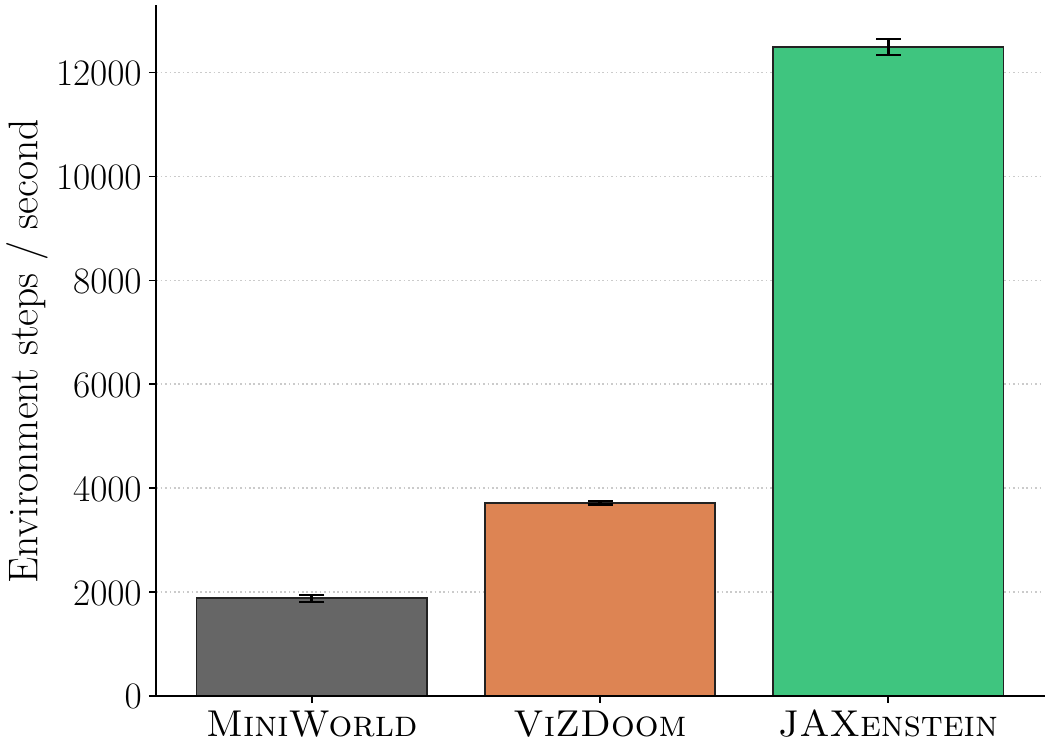}
    \caption{Steps per second without image resizing.}
    \label{fig:env_step_benchmarks_native_ViZDoom_miniworld}
\end{figure}

We also note that MiniWorld and ViZDoom also require an image resizing down to the $64 \times 64$ we require for training a reinforcement learning agent. We plot the same environment step comparison without image resizing in Figure~\ref{fig:env_step_benchmarks_native_ViZDoom_miniworld}. While we do see a modest increase in run time for ViZDoom, the JAXenstein library still achieves an approximately $3\times$ speed up over ViZDoom.

\subsection{Training speed comparison}
\label{appx:train_comparison}

For the training speed comparison, we time the speed it takes different implementations of the recurrent PPO algorithms with different benchmarks to train for 1M steps. For the JAXenstein runs, we run an implementation of recurrent PPO in JAX~\citep{lu2022discovered} on the \texttt{my-way-home} environment. For the ViZDoom runs, we use an implementation of recurrent PPO in PyTorch~\citep{stable-baselines3} with the ViZDoom benchmark on \texttt{VizdoomMyWayHome-v1}. For both implementations, we match network architectures and hyperparameters between the two environments, as well as down-sample the ViZDoom environment frames to $64\times64$. As for the JAXenstein environment, we natively render at $64\times64$. Full implementation details are provided in the open-source baselines codebase.

\section{JAXenstein environment details}
\label{appx:env_details}

\subsection{ASCII maze details}
\label{appx:ascii_maze_details}
Below we define the different characters and what they map to in a JAXenstein environment.

\begin{table}[h]
\centering
\caption{JAXenstein ASCII map symbols.}
\label{tab:ascii-symbols}
\begin{tabular}{ll}
\toprule
Symbol & Meaning \\
\midrule
\texttt{\#} & Default static wall \\
\texttt{1--9}, generated DMLab symbols & Colored static wall \\
\texttt{.} & Floor \\
\texttt{space} & Floor \\
\texttt{S} & Spawn candidate \\
\texttt{G} & Yellow goal candidate \\
\texttt{r} & Red key \\
\texttt{b} & Blue key \\
\texttt{y} & Yellow key \\
\texttt{\char`\"} & Blue unlocked door \\
\texttt{\textbackslash} & Yellow unlocked door \\
\texttt{R} & Red-locked door \\
\texttt{B} & Blue-locked door \\
\texttt{Y} & Yellow-locked door \\
\bottomrule
\end{tabular}
\end{table}

\section{Baseline experiment details}
\label{appx:baseline_results}
For these baseline experiments, we conduct experiments over 5 seeds and use the following set of hyperparameters for PPO:
\begin{table}[ht]
\centering
\caption{PPO hyperparameters.}
\label{tab:ppo-hyperparams}
\begin{tabular}{ll}
\toprule
Hyperparameter & Value \\
\midrule
Rollout steps & 128 \\
PPO epochs & 4 \\
Image size & $64 \times 64$ \\
Learning rate & $2.5 \times 10^{-4}$ \\
GAE $\lambda$ & 0.95 \\
Entropy coefficient & 0.01 \\
Value loss coefficient & 0.5 \\
Max gradient norm & 0.5 \\
GRU hidden size & 256 \\
CNN feature dimension & 256 \\
Previous action concatenation & True \\
Discount factor $\gamma$ & Environment default \\
\bottomrule
\end{tabular}
\end{table}

Environment-specific hyperparameters are listed in the open-source baselines codebase.
We also use and sweep over the following hyperparameters for RND and ICM:

\begin{table}[ht]
\centering
\caption{RND hyperparameters.}
\label{tab:rnd-hyperparams}
\begin{tabular}{ll}
\toprule
Hyperparameter & Value(s) \\
\midrule
RND reward coefficient & $\{0.1, 1.0, 10.0\}$ \\
RND loss coefficient & $\{0.01, 0.1\}$ \\
RND hidden size & 128 \\
RND output size & 128 \\
RND number of layers & 2 \\
Intrinsic advantage coefficient & 1.0 \\
\bottomrule
\end{tabular}
\end{table}

\begin{table}[ht]
\centering
\caption{ICM hyperparameters.}
\label{tab:icm-hyperparams}
\begin{tabular}{ll}
\toprule
Hyperparameter & Value(s) \\
\midrule
ICM learning rate & $\{3 \times 10^{-4}, 1 \times 10^{-4}\}$ \\
ICM reward coefficient & $\{0.1, 1.0, 10.0\}$ \\
ICM latent size & 128 \\
ICM hidden size & 256 \\
ICM update epochs & 1 \\
\bottomrule
\end{tabular}
\end{table}

\end{document}